\documentclass[runningheads]{llncs}
\usepackage{amsmath}
\usepackage{amssymb} 

\usepackage[T1]{fontenc}
\usepackage{graphicx,verbatim}
\usepackage{cite}
\usepackage[colorlinks=true, linkcolor=blue, citecolor=blue, urlcolor=blue]{hyperref}

\newcommand{\equalcontrib}{\textsuperscript{\ensuremath{\dagger}}}
\newcommand{\corrauth}{\textsuperscript{\ensuremath{*}}}

\newcommand{\symbolfootnotetext}[2]{%
  \begingroup
  \renewcommand{\thefootnote}{#1}%
  \footnotetext{#2}%
  \endgroup
}

\usepackage{booktabs, multirow,pifont}
\usepackage{siunitx}
\begin{document}
\title{MedVol-R1: Reward-Driven Evidence Grounding for Volumetric Reasoning Segmentation}
\titlerunning{MedVol-R1}
%

\author{
Zichun Wang\inst{1}\equalcontrib \and
Hairong Shi\inst{2}\equalcontrib \and
Bingzheng Wei\inst{3} \and
Yan Xu\inst{1}\corrauth \and
Zihua Wang\inst{4}\corrauth
}

\authorrunning{Z. Wang et al.}

\institute{
School of Biological Science and Medical Engineering, Beihang University, Beijing, China\\
\email{xuyan04@gmail.com}
\and
Center for Information and Computer Science, School of Science for Open and Environmental Systems, Graduate School of Science and Technology, Keio University, Kanagawa, Japan
\and
Bytedance Inc., China
\and
Tsinghua University, Beijing, China\\
\email{wangzihua07@126.com}
}
  
\maketitle              

\symbolfootnotetext{\ensuremath{\dagger}}{Equal contribution: Zichun Wang and Hairong Shi.}
\symbolfootnotetext{\ensuremath{*}}{Corresponding authors: Yan Xu and Zihua Wang.}

\begin{abstract}
Volumetric Reasoning Segmentation (VRS) aims to segment a target region in a 3D medical scan from a free-form clinical query, where the referent is often implicit and requires both medical knowledge and volume-grounded reasoning.
Existing methods typically rely on specialized segmentation tokens to connect language with mask decoding, but this coupling collapses the decision process into opaque latent representations, limiting interpretability and generalization to diverse narrative expressions.
In this paper, we present \textbf{MedVol-R1}, a reinforcement learning-based framework for VRS that explicitly decouples \emph{evidence grounding} from \emph{volumetric delineation}: the LVLM grounds clinical reasoning to a verifiable 2D evidence anchor (key axial slice and 2D bounding boxes), which is then propagated into a coherent 3D mask by a frozen MedSAM2 model.
We train MedVol-R1 with cold-start supervised fine-tuning followed by GRPO, guided by a multi-component reward that encourages informative evidence selection, accurate 2D spatial grounding, and cross-slice volumetric coherence, without requiring costly chain-of-thought annotations.
Experiments on CT-ORG, AbdomenCT-1K, and KiTS23 from the M3D-Seg benchmark demonstrate that MedVol-R1 consistently outperforms strong baselines and achieves state-of-the-art performance, with reinforcement learning providing clear gains over pure supervised fine-tuning. Code is available \href{https://github.com/AdvancingEther/MedVol-R1}{here}.

\end{abstract}
\section{Introduction}
Volumetric Reasoning Segmentation (VRS) is an important capability for computer-aided diagnosis and clinician--AI interaction~\cite{m3d,segvol, medsurvey1, medsurvey2, medsurvey3}. Given a 3D medical scan and a free-form clinical query, the goal is to produce a voxel-level binary mask for the referred target. Unlike label-based prompts, VRS queries often refer to targets implicitly, requiring either medical world knowledge (e.g., the organ that secretes bile), or volume-grounded reasoning (e.g., the kidney that contains the tumor) to resolve the referent~\cite{m3d}. This requires the model to not only parse language, but also resolve the hidden clinical reference by integrating external medical knowledge with volumetric visual evidence, and then translate this reasoning into reliable 3D localization with spatially coherent voxel masks.

Following the success of reasoning-based visual perception in the general domain~\cite{lisa, lenna, videoespresso}, recent works have extended 3D medical Large Vision Language Models (LVLMs) from coarse semantic interpretation to dense voxel-level segmentation~\cite{vilam3,m3d}. In parallel, prompt-driven and interactive SAM-style medical segmentors have also advanced rapid 2D/3D annotation and refinement~\cite{MedSAM,sammed2d,MedSAM2,hairongmiccai}, but they typically assume an explicit user prompt and do not address implicit narrative reasoning in 3D volumes. Among such LVLM-based methods, a common strategy is to introduce dedicated segmentation tokens (e.g., \texttt{<SEG>}) as latent semantic carriers that connect language-conditioned volumetric representations to a downstream mask decoder for voxel-level prediction. While effective for queries with explicit terminology, this paradigm suffers from two major limitations when confronting implicit, reasoning-dependent clinical narratives. First, by collapsing the entire decision process into implicit token representations, these methods provide no explicit reasoning chain from query interpretation to spatial grounding, offering neither verifiable evidence for target selection nor interpretable rationale for clinical decision support. Second, such imitation-based supervision tends to encourage surface-level pattern mimicry rather than genuine clinical understanding, leading to poor generalization beyond training-time phrasings.

Recent advancements in Group Relative Policy Optimization 
(GRPO)~\cite{deepseekr1}, originally proposed for large 
language model alignment, have been successfully extended 
to vision-language tasks~\cite{segzero, veasonr1, visualrft, medground, medreasoner, medgroundr1}, 
enabling models to discover structured reasoning patterns 
without relying on expensive, manually-annotated Chain-of-Thought (CoT) 
sequences. 
Building on this insight, we propose MedVol-R1, 
the first reinforcement learning-based framework for 
VRS.
MedVol-R1 employs GRPO initialized with a cold-start 
fine-tuning phase to autonomously develop structured 
reasoning capabilities. To further guide the model in the 
complex 3D clinical setting, we design a multi-dimensional 
reward mechanism that balances format constraints, 
hierarchical localization, and cross-slice consistency, 
ensuring segmentation results that are both logically 
grounded and anatomically coherent.

Our primary contributions are summarized as follows:
(1) We propose MedVol-R1, an RL-based framework for VRS that applies GRPO on top of cold-start SFT, without requiring CoT annotations.
(2) We design a multi-component reward that enforces format validity, axial evidence selection, 2D box localization, and cross-slice volumetric consistency.
(3) Extensive experiments on three M3D-Seg CT subsets show state-of-the-art performance.



\begin{figure}[t]
  \centering
  \includegraphics[width=\linewidth]{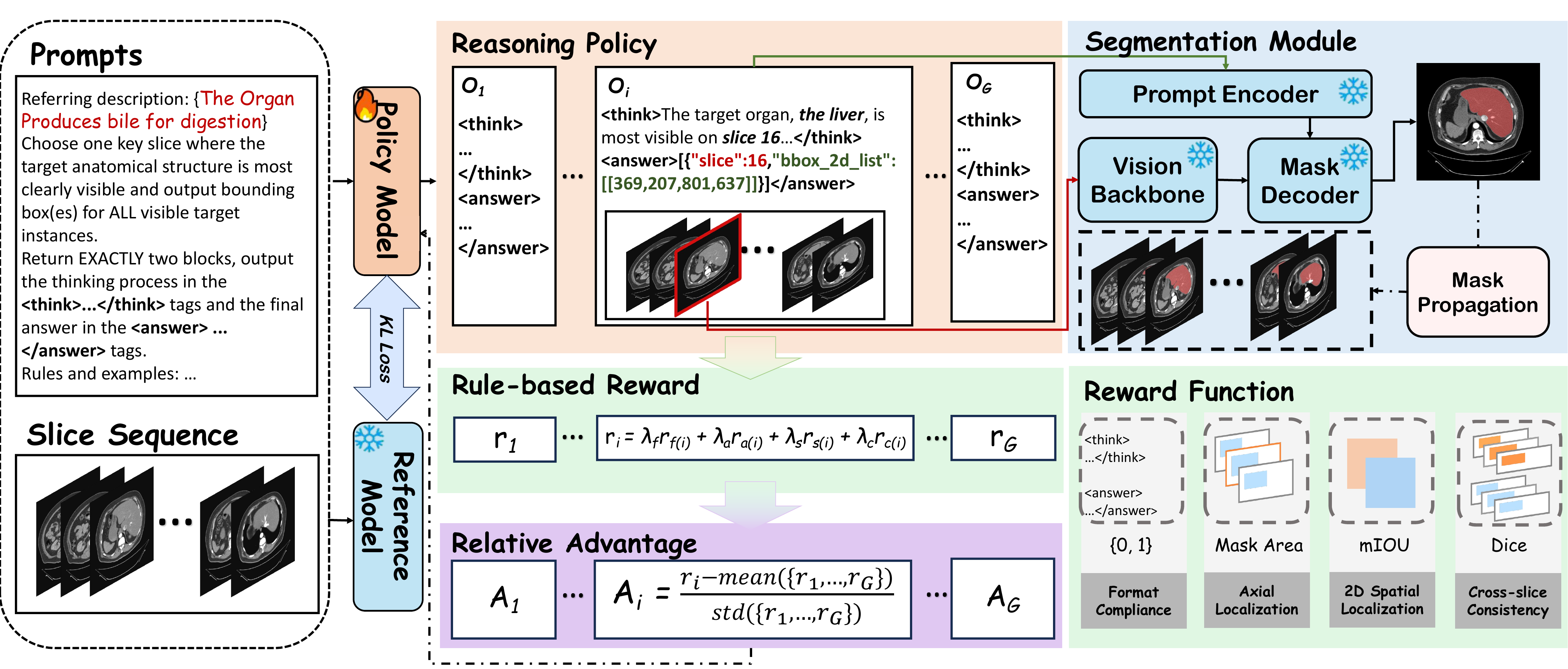}
  \caption{Overall pipeline of MedVol-R1.}
  \label{fig:method}
\end{figure}

\section{Method}

\subsection{Task Formulation}

We consider \textbf{VRS} on CT volumes.
Given a CT volume $\mathbf{V}\in\mathbb{R}^{H\times W\times D}$ and a free-form implicit clinical query $q$,
the objective is to produce a voxel-level binary mask of the referred target:
\begin{equation}
\Phi_\theta:\ (\mathbf{V}, q)\ \mapsto\ \hat{\mathbf{Y}}\in\{0,1\}^{H\times W\times D},
\label{eq:task}
\end{equation}
where $\hat{\mathbf{Y}}$ is the predicted volumetric mask.

Unlike text-prompted volumetric segmentation~\cite{sat,segvol,biomedparsev2}, VRS queries are often implicit, requiring medical knowledge and volume-grounded reasoning to resolve the referent before producing a coherent 3D mask.

\subsection{Preliminary: Group Relative Policy Optimization}
\label{sec:grpo}

Group Relative Policy Optimization \cite{deepseekr1} is a reinforcement learning algorithm that removes the need for an explicit value function by estimating advantages from \emph{relative rewards} within a group of samples. 
Given a prompt $p$, GRPO draws a group of $G$ responses $\{o_i\}_{i=1}^{G}$ from the current policy $\pi_\theta(\cdot\mid p)$. 
A reward function $R(\cdot,\cdot)$ assigns a scalar score to each response, producing
$r_i = R(o_i, p)$ for $i=1,\dots,G$. 
The relative advantage $A_i$ is then computed via group-wise normalization over $\{r_1,\dots,r_G\}$:
\begin{equation}
A_i
=
\frac{
r_i - \mathrm{mean}\!\left(\{r_1, r_2, \dots, r_G\}\right)
}{
\mathrm{std}\!\left(\{r_1, r_2, \dots, r_G\}\right)
}.
\label{eq:grpo_adv}
\end{equation}

GRPO optimizes a clipped policy-gradient objective with a Kullback--Leibler penalty to stabilize training:
\begin{equation}
\mathcal{L}_{\textsc{grpo}}(\theta)
=
-\frac{1}{G}\sum_{i=1}^{G}
\left[
\min\!\left(
\rho_i A_i,\;
\mathrm{clip}\!\left(\rho_i, 1-\epsilon, 1+\epsilon\right) A_i
\right)
-\beta\, D_{\mathrm{KL}}\!\left(\pi_\theta \,\|\, \pi_{\mathrm{ref}}\right)
\right],
\label{eq:grpo_obj}
\end{equation}
where $\rho_i=\frac{\pi_\theta(o_i\mid p)}{\pi_{\theta_{\text{old}}}(o_i\mid p)}$, $\pi_{\mathrm{ref}}$ is the reference policy , and $\beta$ is the regularization coefficient.

\subsection{Pipeline}
Unlike prior approaches~\cite{m3d} that encapsulate 3D semantics 
into opaque specialized tokens, MedVol-R1 explicitly decouples 
volumetric reasoning from geometric delineation.
As illustrated in Fig.~\ref{fig:method}, the framework operates 
in two stages:
(1)~\emph{Evidence Grounding}: given a multi-slice sequence 
$\mathcal{S}$ and a clinical query $q$, the LVLM identifies a 
key slice $I_k$ and predicts 2D bounding boxes 
$\mathcal{B}_k=\{b_i\}_{i=1}^{N_k}$ on that slice;
(2)~\emph{Volumetric Propagation}: a frozen MedSAM2~\cite{MedSAM2} 
lifts this 2D anchor $(I_k,\mathcal{B}_k)$ into a volumetric mask 
$\hat{\mathbf{Y}}$ via slice-to-slice memory propagation.
We optimize the LVLM through cold-start SFT 
(Sec.~\ref{sec:stage1_sft}) followed by GRPO 
(Sec.~\ref{sec:stage2_grpo}).

\subsection{Stage I: Supervised Fine-tuning}
\label{sec:stage1_sft}

We begin with SFT to obtain a \emph{cold-start} policy that mitigates 
the sparse-reward problem in RL~\cite{deepseekr1, segzero, veasonr1}.
Each axial slice in $\mathcal{S}$ is prefixed with a discrete 
identifier (e.g., \texttt{<slice 1>}).
Given a clinical query $q$, the model is trained to predict, within 
\texttt{<answer>} tags, the key-slice index $k$ and the corresponding 
bounding boxes $\mathcal{B}_k$.

Supervision targets are derived from $\mathbf{Y}^*$ by Top-$K$ visibility sampling (rank slices by foreground area, sample one from the Top-$K$), and 2D boxes are obtained by connected-component decomposition on the selected slice mask.

\subsection{Stage II: Reward-Driven Policy Optimization via GRPO}
\label{sec:stage2_grpo}
Building on the SFT-initialized policy, we optimize the LVLM with GRPO~\cite{deepseekr1}.
For each query $q$, the policy $\pi_\theta$ samples a group of $G$ outputs $\{o_i\}_{i=1}^{G}$ under a fixed \texttt{<think>}--\texttt{<answer>} format, where \texttt{<answer>} specifies an evidence anchor (key-slice index $\hat{k}$ and 2D boxes $\mathcal{B}_{\hat{k}}$).
Each output receives a composite reward $r_i=R_{\text{total}}(o_i,q,\mathbf{Y}^*)$ that combines format validity, axial evidence selection, 2D box grounding, and cross-slice consistency.
GRPO updates $\pi_\theta$ using within-group standardized advantages and a KL penalty to a frozen reference model $\pi_{\mathrm{ref}}$~\cite{deepseekr1}.

\paragraph{Reward Function.}
Reward shaping is critical for effective policy optimization under GRPO. 
We therefore design a multi-dimensional reward mechanism $\mathcal{R}$ that evaluates each sampled output from complementary perspectives: 
(i) format compliance for reliable parsing, 
(ii) axial evidence selection for informative key-slice choice, 
(iii) 2D spatial localization for accurate bounding-box grounding, and 
(iv) cross-slice consistency for coherent volumetric delineation under through-plane variations.

\smallskip
\noindent\textbf{Format Compliance Reward.}
$R_f$ enforces a strict output schema:
$R_f\!=\!1$ if the response contains valid 
\texttt{<think>}...\,\texttt{</think>} and 
\texttt{<answer>}...\,\texttt{</answer>} blocks whose 
\texttt{<answer>} content parses into the required schema 
(key-slice index and \texttt{bbox\_2d\_list}), and 
$R_f\!=\!0$ otherwise.

\smallskip
\noindent\textbf{Axial Localization Reward.}
$R_a$ encourages selecting slices where the target is maximally visible.
Letting $F_t$ denote the ground-truth foreground area on slice $t$, we define
$R_a = F_{\hat{k}} / \max_t F_t$,
which provides graded credit proportional to the target visibility on the predicted key slice $\hat{k}$.

\smallskip
\noindent\textbf{2D Spatial Localization Reward.}
$R_s$ measures the spatial precision of predicted bounding boxes on the selected key slice.
Let $\hat{\mathcal{B}}=\{\hat{b}_i\}_{i=1}^{N}$ denote the $N$ predicted boxes and $\mathcal{B}^*=\{b^*_j\}_{j=1}^{N^*}$ the $N^*$ ground-truth boxes on the predicted key slice $I_{\hat{k}}$.
We construct an IoU-based cost matrix $C\in\mathbb{R}^{N\times N^*}$:
\[
C_{i,j}=1-\mathrm{IoU}(\hat{b}_i,b^*_j),
\]
and apply the Hungarian algorithm to obtain an optimal one-to-one matching set $\mathcal{M}$.
The reward is defined as the normalized sum of matched IoUs:
\[
R_s=\frac{1}{\max(N,N^*)}\sum_{(i,j)\in\mathcal{M}}\mathrm{IoU}(\hat{b}_i,b^*_j)\in[0,1],
\]
where the normalization naturally penalizes missed targets and spurious predictions.

\smallskip
\noindent\textbf{Cross-slice Consistency Reward.}
To encourage volumetric coherence beyond a single slice, we introduce a cross-slice consistency reward $R_c$ based on mask propagation.
After obtaining the matched boxes on the predicted key slice (via Hungarian matching), we use them as prompts to an off-the-shelf MedSAM2 model and propagate slice-wise to obtain a volumetric prediction $\hat{\mathbf{Y}}$.
We then compute the Dice score on a local axial neighborhood consisting of the key slice $\hat{k}$ and its adjacent slices:
\[
R_c=\mathrm{Dice}\!\left(\hat{\mathbf{Y}}_{\mathcal{N}(\hat{k})},\mathbf{Y}^*_{\mathcal{N}(\hat{k})}\right)\in[0,1],
\]
where $\mathcal{N}(\hat{k})=\{t~|~|t-\hat{k}|\le \Delta\}$ denotes the set of $\pm\Delta$ adjacent slices around $\hat{k}$, and
$\hat{\mathbf{Y}}_{\mathcal{N}(\hat{k})}$ and $\mathbf{Y}^*_{\mathcal{N}(\hat{k})}$ are the corresponding masks restricted to this neighborhood.
This term directly evaluates whether the predicted evidence anchor can serve as an effective prompt for coherent volumetric delineation, while keeping RL rollouts computationally tractable.

\smallskip
\noindent\textbf{Overall Reward.}
We combine the above components as a weighted sum:
\[
R_{\text{total}} = \lambda_f\, R_f + \lambda_a\, R_a + \lambda_s\, R_s + \lambda_c\, R_c,
\]
where $\lambda_f,\lambda_a,\lambda_s,\lambda_c$ are fixed weights and we set $\lambda_f=\lambda_a=\lambda_s=\lambda_c=1.0$ in all experiments.

\section{Experiments}

\subsection{Experimental Settings}

\subsubsection{Datasets and Evaluation Metrics}
We evaluate our framework on three CT sub-datasets from the M3D-Seg~\cite{m3d} benchmark: CT-ORG, AbdomenCT-1K, and KiTS23.
Following the M3D-Seg protocol, all samples are organized as image--mask--text triplets, where fixed anatomical labels are replaced with free-form linguistic descriptions to support VRS.

To further assess the model's capacity for grounded reasoning over 3D volumes, we paraphrase the original semantic descriptions of all KiTS23 test queries into more diverse expressions.
Unlike fixed category names, these queries couple lesion attributes with anatomical spatial relations (e.g., ``the kidney containing an obvious fluid-filled sac''), requiring the model to jointly interpret semantic cues and 3D structural context.
We report standard voxel-level segmentation metrics, including Dice Similarity Coefficient (DSC) and IoU.

\subsubsection{Compared Methods}
We compare our method with two paradigms of baselines. First, we include two specialized medical parsers, SAT~\cite{sat} and BiomedParseV2~\cite{biomedparsev2}, which are promptable medical segmentation models; we fine-tune them on the M3D-Seg training split using the same free-form referring queries to assess their adaptability to open-ended descriptions. 
Second, we compare with M3D, a native volumetric vision-language baseline for 3D medical image-text understanding.


\subsubsection{Implementation Details}
We instantiate our framework with Qwen3-VL-4B, which provides a practical balance between vision-language reasoning capability and computational feasibility for GRPO-based volumetric evidence grounding.
Each 3D volume is represented as a uniformly sampled sequence of 64 axial slices resized to $256\times256$.
Training follows a two-stage pipeline.
\textbf{(i) SFT.} We fine-tune the model for 1 epoch using LoRA (rank 128, all-linear) with a learning rate of $2\times10^{-5}$ and a per-GPU batch size of 2.
\textbf{(ii) RL.} We then optimize the policy for 3 epochs with GRPO ($G{=}4$, $\beta{=}0.01$), using a learning rate of $2\times10^{-6}$ and a per-GPU batch size of 1.
Both stages use AdamW with a 0.1 warmup ratio and a cosine learning-rate schedule.
All experiments are run on NVIDIA RTX A6000 GPUs with bfloat16 precision and gradient checkpointing.
The maximum generation length is 256 tokens.


\begin{table}[t]
  \centering
  \caption{Quantitative comparison with state-of-the-art methods.}
  \label{tab:main}
  \setlength{\tabcolsep}{4pt}
  \renewcommand{\arraystretch}{1.15}

  \begin{tabular}{l
    S[table-format=2.2] S[table-format=2.2]
    S[table-format=2.2] S[table-format=2.2]
    S[table-format=2.2] S[table-format=2.2]}
    \toprule
    \multirow{2}{*}{\textbf{Method}} &
    \multicolumn{2}{c}{\textbf{AbdomenCT-1K}} &
    \multicolumn{2}{c}{\textbf{CT-ORG}} &
    \multicolumn{2}{c}{\textbf{KiTS23}} \\
    \cmidrule(lr){2-3}\cmidrule(lr){4-5}\cmidrule(lr){6-7}
    & {\textbf{DSC}} & {\textbf{IoU}}
    & {\textbf{DSC}} & {\textbf{IoU}}
    & {\textbf{DSC}} & {\textbf{IoU}} \\
    \midrule
    SAT~\cite{sat}                & 58.44 & 41.28 & 48.64 & 32.14 & 28.70 & 16.75 \\
    BiomedParseV2~\cite{biomedparsev2}       & 68.78 & 52.42 & 40.23 & 25.18 & 32.94 & 19.72 \\
    M3D~\cite{m3d}                & 73.63 & 58.27 & 83.49 & 71.66 & 30.69 & 18.13 \\
    \midrule
    Ours (Pure SFT) & 85.52 & 74.70 & 83.23 & 71.28 & 36.21 & 22.11 \\
    Ours (SFT+RL)   & \textbf{89.86} & \textbf{81.59} & \textbf{85.43} & \textbf{74.57} & \textbf{45.46} & \textbf{29.42} \\
    \bottomrule
  \end{tabular}
\end{table}

\subsection{Comparisons with Other Methods}

Table~\ref{tab:main} compares MedVol-R1 with representative baselines
on AbdomenCT-1K, CT-ORG, and KiTS23.
Our full model (SFT+RL) achieves the highest DSC and IoU on all three
benchmarks, with particularly large margins on AbdomenCT-1K
(89.86 vs.\ 73.63 DSC) and KiTS23 (45.46 vs.\ 30.69 DSC).
Among the baselines, M3D remains competitive on CT-ORG where queries
use standard terminology, but degrades sharply on KiTS23 whose
paraphrased queries require joint interpretation of lesion attributes
and spatial context.
SAT and BiomedParseV2, designed for explicit category prompts,
fail to generalize to free-form narrative queries despite fine-tuning.
GRPO brings consistent gains over pure SFT (+4.34 DSC on AbdomenCT-1K,
+2.20 on CT-ORG, +9.25 on KiTS23), with the largest improvement on
the most reasoning-heavy subset, confirming that reward-driven
optimization is especially effective when queries demand
non-trivial clinical inference.

\begin{figure}[t]
  \centering
  \includegraphics[width=\linewidth]{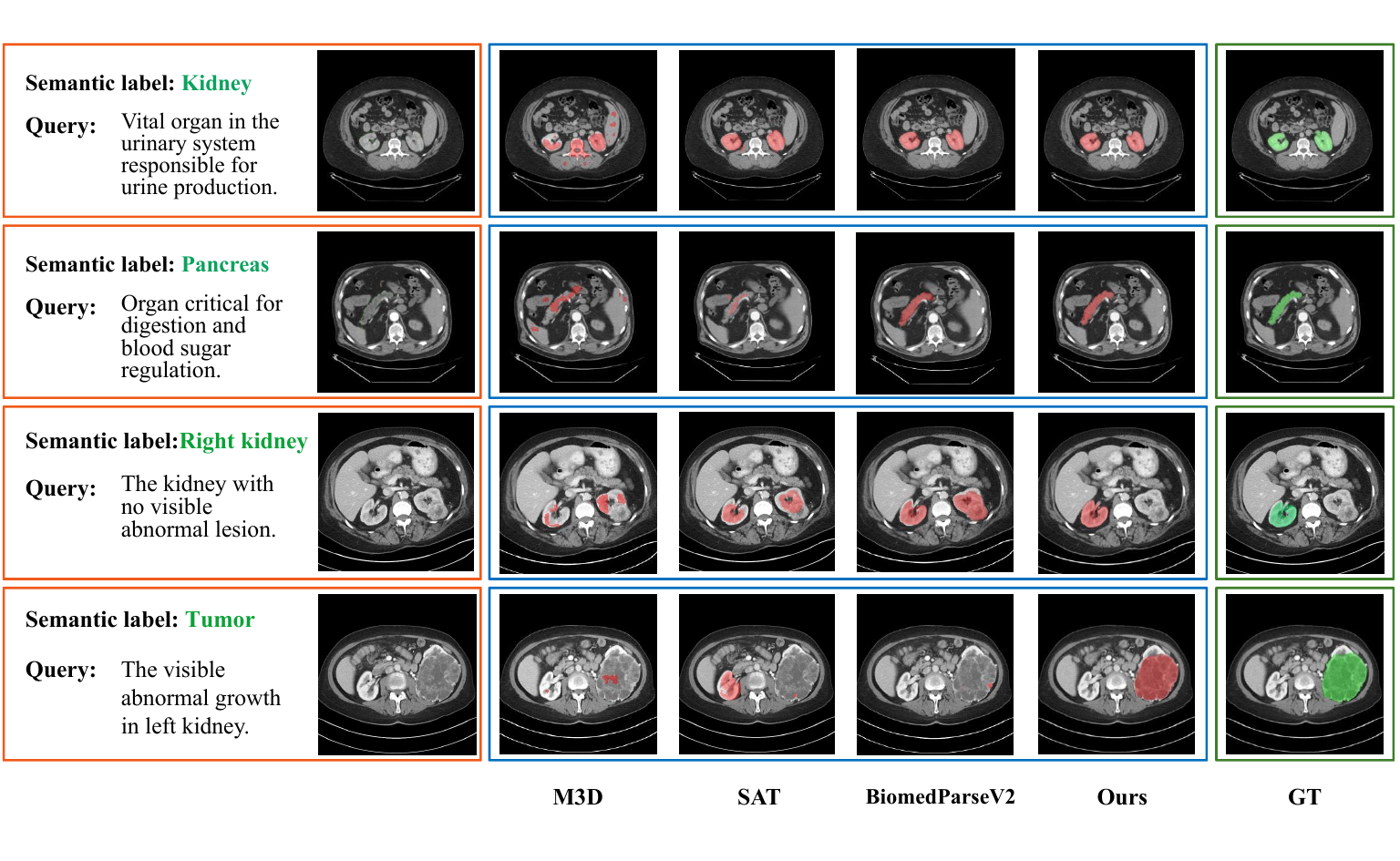}
  \caption{Qualitative comparisons on four representative VRS samples.}
  \label{fig:qual}
\end{figure}

\subsection{Qualitative Results}
Fig.~\ref{fig:qual} presents qualitative comparisons on four representative samples.
Across all cases, MedVol-R1 produces segmentation masks that more closely match the ground-truth (GT) than the strongest baseline M3D.
In particular, our results show cleaner boundaries and reduced leakage to adjacent structures, while preserving more complete target coverage.
This indicates that the RL-enhanced policy provides more reliable evidence anchors for downstream MedSAM2 propagation, consistent with the quantitative gains in Table~\ref{tab:main}.

\begin{table}[t]
\centering
\footnotesize
\renewcommand{\arraystretch}{1.12}

\begin{minipage}[t]{0.60\linewidth}
\centering
\setlength{\tabcolsep}{5.5pt}
\textbf{(a) Effect of Reward Components}\\[-0.15em]
\begin{tabular}{@{}lcc@{}}
\toprule
\textbf{Setting} & \textbf{DSC} & \textbf{IoU} \\
\midrule
RL only (w/o SFT) & -- & -- \\
SFT only          & 85.52 & 74.70 \\
w/o $R_c$         & 88.61 & 79.55 \\
w/o $R_s$         & 87.58 & 77.90 \\
w/o $R_a$         & 89.17 & 80.46 \\
w/o $R_f$         & 88.98 & 80.53 \\
Full              & \textbf{89.86} & \textbf{81.59} \\
\bottomrule
\end{tabular}
\end{minipage}
\hfill
\begin{minipage}[t]{0.34\linewidth}
\centering
\vspace{1.55em}
\setlength{\tabcolsep}{8pt}
\textbf{(b) Effect of $\Delta$}\\[-0.15em]
\begin{tabular}{@{}ccc@{}}
\toprule
\textbf{$\Delta$} & \textbf{DSC} & \textbf{IoU} \\
\midrule
2  & 84.41 & 73.03 \\
5  & \textbf{85.43} & \textbf{74.57} \\
10 & 85.16 & 74.16 \\
15 & 83.99 & 72.40 \\
\bottomrule
\end{tabular}
\end{minipage}

\vspace{0.35em}
\caption{\textbf{Ablation studies.}
(a) Effect of reward components on AbdomenCT-1K. ``w/o'' denotes removing one reward term from the full GRPO objective.
(b) Effect of the local axial window size $\Delta$ used in the cross-slice consistency reward $R_c$ on CT-ORG.}
\label{tab:ablation}
\end{table}

\subsection{Ablation Study}
Table~\ref{tab:ablation}(a) ablates key design choices of our 
training pipeline on AbdomenCT-1K.
RL without SFT warm-start fails to converge, confirming that 
cold-start initialization is essential for stable policy optimization.
Compared with SFT-only, the full model improves DSC from 85.52 to 89.86, 
showing the benefit of GRPO over pure imitation learning.
Removing any single reward component degrades performance, validating 
their complementarity.
Among them, removing $R_s$ causes the largest decline 
(DSC 87.58, IoU 77.90), indicating that accurate 2D spatial grounding 
is critical for downstream volumetric propagation.
Removing $R_c$ yields the second-largest drop, confirming the importance 
of cross-slice consistency.

Table~\ref{tab:ablation}(b) studies the neighborhood size $\Delta$ in 
$R_c$ on CT-ORG.
$\Delta=5$ achieves the best performance, while smaller windows provide 
limited volumetric supervision and larger windows may introduce noisy 
signals from distant slices with weak or ambiguous target visibility.
Thus, we use $\Delta=5$ by default.




\section{Conclusion}
We presented MedVol-R1, a reinforcement learning-based framework
for Volumetric Reasoning Segmentation that explicitly decouples
evidence grounding from volumetric delineation.
The LVLM was trained to predict a verifiable 2D evidence anchor
(key axial slice and bounding boxes), which a frozen MedSAM2
propagated into a coherent 3D mask.
A two-stage pipeline of cold-start SFT followed by GRPO,
guided by a multi-component reward covering format validity,
axial evidence selection, spatial grounding, and cross-slice
consistency, avoids the need for manually annotated chain-of-thought supervision.
Experiments on three M3D-Seg CT subsets showed consistent
improvements over strong baselines, with GRPO providing clear
gains beyond pure supervised fine-tuning.
Future work will extend MedVol-R1 to multi-modal volumetric data, such as MRI and ultrasound volumes and explore richer evidence representations
beyond single-slice anchors to further improve volumetric coherence.

\subsubsection*{Acknowledgements}
This work was supported by the National Natural Science Foundation of China under Grants 62371016 and U23B2063, the Fundamental Research Funds for the Central Universities from the State Key Laboratory of Software Development Environment at Beihang University, the 111 Project under Grant B13003, and the high-performance computing (HPC) resources at Beihang University.

\subsubsection{Disclosure of Interests.}
We have no conflicts of interest to disclose.
%
%
%
%
\bibliographystyle{splncs04}
\bibliography{Paper-2593}

\end{document}